\documentclass[10pt,twocolumn,letterpaper]{article}

\usepackage[pagenumbers]{cvpr} % To force page numbers, e.g. for an arXiv version

\definecolor{cvprblue}{rgb}{0.21,0.49,0.74}
\usepackage[pagebackref,breaklinks,colorlinks,allcolors=cvprblue]{hyperref}
\usepackage{booktabs}
\usepackage{colortbl}
\usepackage{xcolor}
\usepackage[normalem]{ulem}
\useunder{\uline}{\ul}{}
\usepackage{float}
\usepackage{multirow}
\usepackage{amssymb,mathrsfs,amsmath}

\title{PO3AD: Predicting Point Offsets toward Better 3D Point Cloud Anomaly Detection}

\author{
    Jianan Ye\textsuperscript{\rm 1}$^{,}$\textsuperscript{\rm 2}$^{, \ast}$,
    Weiguang Zhao\textsuperscript{\rm 1}$^{,}$\textsuperscript{\rm 2}$^{,}$\thanks{Equal contribution.} ,
    Xi Yang\textsuperscript{\rm 1},
    Guangliang Cheng\textsuperscript{\rm 2},
    Kaizhu Huang\textsuperscript{\rm 3}\thanks{Corresponding author.}\\
    \textsuperscript{\rm 1} School of Advanced Technology, Xi’an Jiaotong-Liverpool University\\
    \textsuperscript{\rm 2} School of Electrical Engineering, Electronics and Computer Science, University of Liverpool\\
    \textsuperscript{\rm 3} Data Science Research Center, Duke Kunshan University
}

\begin{document}
\maketitle

\begin{abstract}
Point cloud anomaly detection under the anomaly-free setting poses significant challenges as it requires accurately capturing the features of 3D normal data to identify deviations indicative of anomalies. Current efforts focus on devising reconstruction tasks, such as acquiring normal data representations by restoring normal samples from altered, pseudo-anomalous counterparts. Our findings reveal that distributing attention equally across normal and pseudo-anomalous data tends to dilute the model's focus on anomalous deviations. The challenge is further compounded by the inherently disordered and sparse nature of 3D point cloud data. In response to those predicaments, we introduce an innovative approach that emphasizes learning \textit{point offsets}, targeting more informative pseudo-abnormal points, thus fostering more effective distillation of normal data representations. We also have crafted an augmentation technique that is steered by \textit{normal vectors}, facilitating the creation of credible pseudo anomalies that enhance the efficiency of the training process. Our comprehensive experimental evaluation on the Anomaly-ShapeNet and Real3D-AD datasets evidences that our proposed method outperforms existing state-of-the-art approaches, achieving an average enhancement of 9.0\% and 1.4\% in the AUC-ROC detection metric across these datasets, respectively.
\end{abstract}

\section{Introduction}

Point cloud anomaly detection aims to identify defective samples and locate abnormal regions that deviate from expected data patterns~\cite{roth2022towards, zhou2024r3dad}. Owing to the high cost of collecting and labeling anomaly samples, this task is usually implemented in an anomaly-free setting, i.e., only normal samples are available during training.
The critical challenge within this framework is to effectively capture the distinctive features that are characteristic of 3D normal data, enabling the system to recognize and classify instances that deviate from these normal patterns as anomalies.
Moreover, the inherently disordered and sparse nature of 3D point cloud data significantly complicates the process of acquiring such discriminative knowledge.

\begin{figure}
    \centering
    \includegraphics[width=\linewidth]{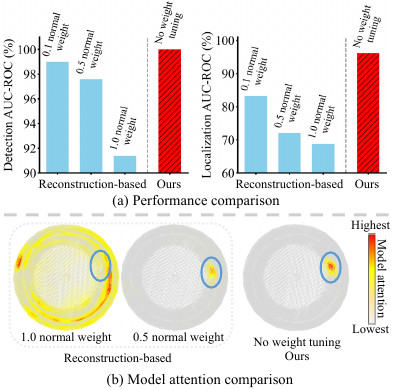}
    \caption{Comparison of reconstruction-based method and our method in terms of performance, and model attention. (a) Detection and localization performance of the reconstruction-based method on the ashtray0 category with various normal point loss weights; pseudo-abnormal points consistently weighted at 1.0 (implemented with our network due to the absence of official code). (b)By reducing normal weight in the reconstruction-based method, the model pays more attention to pseudo-abnormal points (marked with blue circles). Our method successfully focuses on pseudo-abnormal points. The model attention map is obtained by calculating the gradient of each point during backward propagation.}
    \label{fig:banner_motivation}
\end{figure}

\begin{figure}
    \centering
    \includegraphics[width=\linewidth]{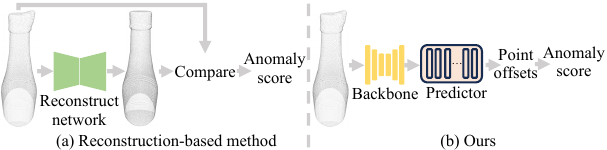}
    \caption{Model structure comparison. (a) Restores normal samples from pseudo-abnormal variants; anomaly scores from input-output comparison. (b) Predicts point offsets of pseudo anomalies; anomaly scores from predicted offsets during testing.}
    \label{fig:banner_framework}
\end{figure}

As one reasonable way to tackle this task, recent efforts focus on designing reconstruction tasks to capture normal representations, as illustrated in Figure~\ref{fig:banner_framework}\textcolor{cvprblue}{(a)}.  For instance, IMRNet~\citep{li2024towards} detects anomalies by reconstructing randomly masked normal point cloud samples and comparing inputs with their reconstruction outputs.
This approach may fail to detect anomalies in unmasked regions. To address this limitation, R3D-AD~\citep{zhou2024r3dad} reconstructs normal samples from their pseudo-abnormal variants. However, reconstructing each point's coordinates in 3D space causes equal loss weights for both normal and pseudo-abnormal points. Extraction of normal patterns relies on learning to restore normal regions from pseudo-abnormal ones, but equal loss weights impair the network to focus on this process, thus limiting the detection performance. Empirical evidence in Figure~\ref{fig:banner_motivation} shows that the model focuses more on pseudo-abnormal regions as normal point loss weight decreases (the loss weight of pseudo-abnormal points is fixed at 1.0), leading to a performance boost.

In this paper, we propose to predict \textit{point offsets} in pseudo anomalies (as illustrated in Figure~\ref{fig:banner_framework}\textcolor{cvprblue}{(b)}), which allows the model to concentrate on pseudo-abnormal regions and thus effectively distil normal representations.
Point offsets are essentially vectors characterized by two attributes: magnitude and direction.
The offsets of abnormal points in pseudo-anomalies are defined by these attributes, representing their displacement distance and direction relative to their corresponding points in original normal ones.
In contrast, the offsets of normal points in pseudo anomalies can be predominantly governed by their displacement distance, as they remain unchanged relative to their corresponding points in original normal ones, making the direction less relevant and the magnitude zero.
Learning the task of point offset prediction allows the model to estimate normal points' offset magnitude only, while requiring it to predict both offset magnitude and direction for pseudo-anomaly points. This approach significantly diverges from the current mainstream reconstruction-based methods, which require precise restoration of point coordinates and consequently distribute focus equally among normal and pseudo-abnormal points.
Empirical evidence supporting our method is presented in Figure~\ref{fig:banner_motivation}\textcolor{cvprblue}{(b)}.
On the right, our method successfully focuses on pseudo-abnormal regions, whereas the reconstruction-based method (1.0 normal weight) fails to do so.
Additionally,  during inference, predicted offsets serve as direct indicators of abnormality levels, while reconstruction-based methods rely on manually designed metrics to produce anomaly scores.

Drawing inspiration from the aforementioned observation, we propose a novel framework named PO3AD, which efficiently predicts point offsets and adequately captures normal representations.
For practical implementation, in order to enable the model to learn the knowledge of predicting offsets, we further propose an anomaly simulation method named Norm-AS, which is guided by \textit{normal vectors}\footnote{In this paper, `\textit{normal vectors}' exclusively refers to the vectors perpendicular to the surface in point cloud geometry, while `normal' denotes non-abnormal. To avoid confusion, we italicized \textit{normal vectors}.}.
Our Norm-AS leverages \textit{normal vectors} to control point movement direction, enabling the creation of credible pseudo anomalies that resemble real ones (as shown in Figure~\ref{fig:pseudobanner}\textcolor{cvprblue}{(d)}), thus increasing learning efficiency.
In contrast, the previous augmentation method~\cite{zhou2024r3dad} ignores point movement direction. This may cause pseudo-abnormal regions to overlap with normal regions (as shown in Figure~\ref{fig:pseudobanner}\textcolor{cvprblue}{(c)}), which consequently confuses the model, leading to less effective learning.
The offsets of points in pseudo anomaly samples relative to their original normal counterparts serve as training labels.
During testing, the predicted offsets are used to recognize anomalies.

Our contributions can be summarized as follows:
\begin{itemize}
    \item A novel paradigm named PO3AD is proposed to predict point offsets, allowing the model to concentrate on pseudo-abnormal regions and ensuring the effective learning of normal representations for 3D point cloud anomaly detection.
    \item  A point cloud pseudo anomaly generation method guided by \textit{normal vectors}, termed Norm-AS, is designed to create credible pseudo anomalies from normal samples for improving training efficiency.
    \item Extensive experiments conducted on two benchmark point cloud anomaly detection datasets demonstrate the superiority of our method to state-of-the-art methods, with an average improvement of 9.0\% and 1.4\% detection AUC-ROC on Anomaly-ShapeNet and Real3D-AD, respectively.
\end{itemize}

\section{Related Work}
\textbf{2D anomaly detection.}
Anomaly detection methods on 2D image data under anomaly-free scenarios have been widely studied in recent years.
To address the issue that anomalies are unavailable during training, a straightforward approach involves generating pseudo anomalies~\cite{hu2024anomalydiffusion, zavrtanik2021draem, li2021cutpaste, schluter2022natural, liu2023simplenet, zhang2024realnet}, allowing models to learn discriminative knowledge for identifying anomalies.
An alternative way to tackle this task relies on constructing a memory bank storing normal features produced by pre-trained encoders~\cite{bae2023pni, kim2023fapm, roth2022towards, xie2023pushing}. Such methods detect anomalies by contrasting features of test data with those of normal training samples.
Flow-based methods~\cite{rudolph2021same, gudovskiy2022cflow} leverage normalizing flows for estimation of the feature distribution to detect anomalies.
Reconstruction-based methods~\cite{huang2022self, pirnay2022inpainting, yan2021learning, zavrtanik2021reconstruction} designs reconstruction tasks to capture normal representations; anomalies are detected by comparing inputs to their reconstruction results.
In this paper, we focus on 3D point cloud anomaly detection. This task is particularly challenging due to the disordered and sparse characteristics of point cloud data.

\begin{figure}[!t]
    \centering
    \includegraphics[width=\linewidth]{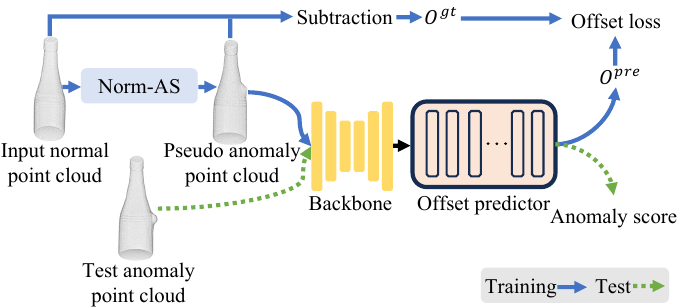}
    \caption{Illustration of our framework. Norm-AS generates pseudo anomalies from training normal samples. The backbone extracts features from pseudo anomalies, and the offset predictor estimates offsets for each point of input. The network trains under an offset loss constraint. During inference, the predicted offset distances serve as anomaly scores for test instances.
    }
    \label{fig:framework}
\end{figure}

\textbf{3D anomaly detection.}
Although significant progress has been made in 2D anomaly detection, research into anomaly detection for 3D data is still relatively limited.
Early studies~\cite{rudolph2023asymmetric,wang2023multimodal,cao2024complementary} focus on combining 3D geometric information with 2D image features to tackle this task. In contrast, 3D-ST~\cite{bergmann2023anomaly} and BTF~\cite{horwitz2023back} detect anomalies by extracting only 3D features through 3D descriptors and have demonstrated effective performance.
With the proposal of two pure point cloud anomaly detection datasets: Real3D-AD~\cite{liu2023real3d} and Anomaly-ShapeNet~\cite{li2024towards}, recent efforts focus on anomaly detection for pure point cloud data.
Reg3D-AD~\cite{liu2023real3d} and Group3AD~\cite{zhu2024towards} combine the classical 2D method PatchCore~\cite{roth2022towards} with RANSAC algorithm~\cite{bolles1981ransac} to develop memory bank-based frameworks for point cloud anomaly detection.
Despite their effectiveness, they suffer the prohibitive computational and storage.
IMRNet~\cite{li2024towards} and R3D-AD~\cite{liu2023simplenet} adopt 2D reconstruction-based methods. They train models to restore normal data from masked or pseudo-abnormal variants, and then detect anomalies by comparing inputs with outputs during the test.
Unlike previous methods, we make a first attempt and propose to predict point offsets to capture effective normal point cloud representations for anomaly detection.

\textbf{Point offset learning.} Predicting point offsets has been widely applied in 3D segmentation and object detection~\cite{jiang2020pointgroup,zhao2023divide,he2021dyco3d,vu2022softgroup}. Yet, it has not been explored in 3D anomaly detection. We make a first effort to apply point offset learning for point cloud anomaly detection, promoting the model to focus on pseudo-abnormal points and thereby enabling effective extraction of normal representations.

\section{Methodology}
\textbf{Problem statement.} Point cloud anomaly detection involves a training set $\mathcal{D}_{train}^e=\{P_q \in \mathbb{R}^{N\times3}\}_{q=1}^{M}$, which consists of $M$ normal samples with $N$ points, belonging to a specific category $e$. A test set, $\mathcal{D}_{test}^e=\{P_q \in \mathbb{R}^{N\times3}, t_q \in \mathcal{T}\}_{q=1}^K$, consists of samples $P_q$ with labels $t_q$, where $\mathcal{T}=\left\{0,1\right\}$ (0 denotes a normal and 1 denotes an anomaly). The objective is to train a deep anomaly detection model on $\mathcal{D}_{train}^e$ to build a scoring function $\phi$: $\mathbb{R}^{N\times3} \rightarrow \mathbb{R}$ that quantitatively evaluate the abnormality levels of new point cloud instances.

\textbf{Overview.} The overview of our framework is presented in Figure~\ref{fig:framework}. 
Given one sample for illustrating our procedure, a pseudo anomaly point cloud is generated from it by our Norm-AS. The subtraction of the input normal sample from the pseudo-abnormal one is used as the training label. Then, the pseudo anomaly is fed into a backbone to extract its features. An offset prediction module then takes these features as input to produce the prediction results. Afterward, the model parameters are optimized by an offset loss. During testing, the predicted offsets are applied to test data to evaluate their abnormal levels.

\subsection{Offset Prediction Learning}
To capture normal representation for anomaly detection, we propose to predict point offsets. Practically, we construct an offset prediction network and leverage an offset loss to supervise the network in learning the knowledge of estimating points offsets.
Our offset prediction learning allows the model to concentrate on pseudo-abnormal points, as evidenced in Figure~\ref{fig:banner_motivation}\textcolor{cvprblue}{(b)}. This enables effective normal representation extraction for anomaly detection.

\subsubsection{Offset Prediction Network}
Our network is composed of two modules: a backbone and an offset predictor.
Inspired by exemplary pioneering work~\cite{hu2021bidirectional,zhao2023divide,Schult23ICRA,delitzas2024scenefun3d} in 3D domain, we adopt MinkUNet~\cite{choy2019fully,choy20194d} as the backbone for our method.
Specifically, MinkUNet is a voxel-based sparse convolutional network~\cite{Graham2015Sparse3C,gwak2020gsdn}  that effectively captures detailed local features from point clouds. This allows the extraction of fine-grained pseudo-abnormal features during training, thus facilitating normal representation learning.
Given one point cloud sample $P\in \mathbb{R}^{N\times3}$, it is voxelized into $V \in \mathbb{R}^{N_V\times3}$, where $N_V$ stands for the number of voxels. It is noted that $N_V \leq N$ and $N_V$ are inversely correlated with the voxel size. The MinkUNet $f_U$ maps $V$ to latent voxelized features $G^V \in \mathbb{R}^{N_V \times C} = f_U(V)$, where $C$ denotes the dimension of each voxel's feature. Then, the voxel-to-point index is leveraged to transform $G^V$ to latent point features $G^P \in \mathbb{R}^{N \times C}$, which are utilized to predict point-wise offsets.
Our offset predictor $f_O$ is built using a Multi-Layer Perceptron (MLP), which takes $G^P$ as input to estimate the offset of each point $O^{pre} \in \mathbb{R}^{N \times 3} = f_O(G^P)$. The offset of each point is composed of three coordinate (xyz) offsets. Each element in $O^{pre}$ refers to the offset of a point along a particular coordinate. $O^{gt} \in \mathbb{R}^{N \times 3}$ is obtained by performing $\hat{P} - P$, where $\hat{P}$ is a pseudo anomaly sample created from $P$ through the Norm-AS.

\subsubsection{Offset Loss}
An offset loss is adopted to guide the network in learning the knowledge of predicting point offsets.
These point offsets are vectors that describe the displacement distance and direction of each point in pseudo anomalies compared to its corresponding point in normal ones.
Accordingly, an L1 loss and a negative cosine loss are employed to supervise the network in predicting point offset distance and direction, respectively, yielding an offset loss:
\begin{align}
    &\mathcal{L}_{off} = \mathcal{L}_{dist} + \mathcal{L}_{dir}, \\
    &\mathcal{L}_{dist} = \frac{1}{N} \sum_{i=1}^N \left\|o_i^{pre} - o_i^{gt} \right\|_{o_i^{pre} \in O^{pre}, o_i^{gt} \in O^{gt}}, \\
    &\mathcal{L}_{dir} = -\frac{1}{N} \sum_{i=1}^N {\frac{o_i^{pre}}{\|o_i^{pre}\|_2 + \epsilon} \cdot \frac{o_i^{gt}}{\|o_i^{gt}\|_2 + \epsilon}} \Big|_{\substack{o_i^{pre} \in O^{pre}\\o_i^{gt} \in O^{gt}}},
\end{align}
where $\mathcal{L}_{dist}$ and $\mathcal{L}_{dir}$  are equally weighted to avoid a possible bias to one loss. Here, $\epsilon$ is set to 1e-8 to prevent division by zero.
It is worth noting that $L_{dir}$ works for pseudo-abnormal points only since the ground truth offset for each normal point is a zero vector.
The significance of $\mathcal{L}_{dist}$ and $\mathcal{L}_{dir}$ in capturing normal representations is demonstrated in Section~\ref{sec:ablation}.

\subsection{Norm-AS}

To create credible pseudo anomalies to improve training efficiency, we develop a novel anomaly simulation method guided by \textit{normal vectors}.
Our proposed Norm-AS is performed by moving the points of a random region along the \textit{normal vectors} or in the opposite direction, generating anomaly types of bulge or concavity.
The region is selected by dividing a point cloud into multiple patches and then randomly sampling one of these patches.
Given a training normal point cloud sample $P \in \mathbb{R}^{N\times3}$, it is divided into $J$ patches as $PH = \{{ph}_b \in \mathbb{R}^{N_h \times3}\}_{b=1}^J$, where $N_h$ is the number of points in each patch and is equal to $N / J$.
Specifically, each patch is determined iteratively by randomly selecting one point and its nearest $N_h - 1$ points from $P^r$.
$P^r$ denotes the points in the point cloud  $P$ that have not been included in any patches.
Based on this, $ph_b$ exhibits various shapes rather than being only circular, enabling the creation of pseudo anomalies with various shapes. A randomly sampled $ph_b$ is then produced as a pseudo-abnormal region by:
\begin{equation}
    \hat{ph_b} = ph_b + \alpha \cdot nv_b \cdot (1-w) \cdot \beta,
\end{equation}
where $nv_b  \in \mathbb{R}^{N_h \times3}$ is the \textit{normal vectors} of $ph_b$. $\alpha$ is randomly sampled from $\{-1, 1\}$ to control whether the point moves along the $nv_b$ ($\alpha = 1$) or in the opposite direction ($\alpha = -1$).
$w$ refers to a matrix with $N_h$ elements, each representing the normalized distance of a point in $ph_b$ from the center point. By performing $1-w$, we aim to move the center point the greatest distance, while points farther from the center are moved shorter distances.
$\beta$ denotes the movement distance of the center point. It is sampled from a uniform distribution with the empirically set range $[0.06, 0.12]$ to produce pseudo anomalies with varying offset distances.
A pseudo anomaly is produced by replacing the corresponding region in $P$ with $\hat{ph_b}$.
The size of the pseudo-abnormal region is determined by $J$, and its impact on normal representation learning is discussed in Section~\ref{sec:sen}.

\begin{figure}
    \centering
    \includegraphics[width=\linewidth]{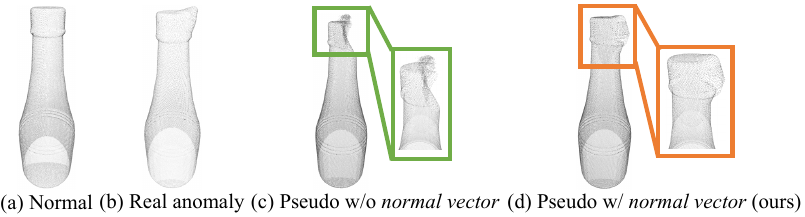}
    \caption{Visualization of pseudo samples with and without \textit{normal vectors} on the bottle0 category. Samples generated with \textit{normal vectors} better mimic real anomalies.}
    \label{fig:pseudobanner}
\end{figure}

The Norm-AS enables the creation of pseudo anomalies resembling real ones, as evidenced in Figure~\ref{fig:pseudobanner}\textcolor{cvprblue}{(d)}.
In contrast, pseudo anomalies generated without the guidance of \textit{normal vectors}, shown in Figure~\ref{fig:pseudobanner}\textcolor{cvprblue}{(c)}, contain pseudo-abnormal points that overlap with normal ones. This may hinder the model from extracting effective features of this region, thus reducing training efficiency.
More examples of our pseudo anomalies are provided in Figure~\ref{fig:pseudovis} of Supplementary~\ref{sec:pseudovis}.
The significance of generating pseudo anomalies guided by \textit{normal vectors} for normal representation learning is validated in Section~\ref{sec:ablation}.

\begin{table*}[!t]
\centering
\resizebox{\linewidth}{!}{
\begin{tabular}{l|cccccccccccccc}
\toprule
Method                      & ashtray0       & bag0          & bottle0       & bottle1       & bottle3       & bowl0         & bowl1         & bowl2         & bowl3         & bowl4         & bowl5         & bucket0       & bucket1       & cap0          \\ \midrule
BTF (Raw) (CVPR 23')        & 57.8           & 41.0          & 59.7          & 51.0          & 56.8          & 56.4          & 26.4          & 52.5          & 38.5          & 66.4          & 41.7          & 61.7          & 32.1          & 66.8          \\
BTF (FPFH)                  & 42.0           & 54.6          & 34.4          & 54.6          & 32.2          & 50.9          & 66.8          & 51.0          & 49.0          & 60.9          & 69.9          & 40.1          & 63.3          & 61.8          \\
M3DM (CVPR 23')             & 57.7           & 53.7          & 57.4          & 63.7          & 54.1          & 63.4          & 66.3          & 68.4          & 61.7          & 46.4          & 40.9          & 30.9          & 50.1          & 55.7          \\
PatchCore (FPFH) (CVPR 22') & 58.7           & 57.1          & 60.4          & 66.7          & 57.2          & 50.4          & 63.9          & 61.5          & 53.7          & 49.4          & 55.8          & 46.9          & 55.1          & 58.0          \\
PatchCore (PointMAE)        & 59.1           & 60.1          & 51.3          & 60.1          & 65.0          & 52.3          & 62.9          & 45.8          & 57.9          & 50.1          & 59.3          & 59.3          & 56.1          & 58.9          \\
CPMF (PR 24')               & 35.3           & 64.3          & 52.0          & 48.2          & 40.5          & 78.3          & 63.9          & 62.5          & 65.8          & 68.3          & 68.5          & 48.2          & 60.1          & 60.1          \\
Reg3D-AD (NeurIPS 23')      & 59.7           & 70.6          & 48.6          & 69.5          & 52.5          & 67.1          & 52.5          & 49.0          & 34.8          & 66.3          & 59.3          & 61.0          & 75.2          & 69.3          \\
IMRNet (CVPR 24')           & 67.1           & 66.0          & 55.2          & 70.0          & 64.0          & 68.1          & 70.2          & 68.5          & 59.9          & 67.6          & {\ul 71.0}    & 58.0          & {\ul 77.1}    & 73.7          \\
R3D-AD (ECCV 24')           & {\ul 83.3}     & {\ul 72.0}    & {\ul 73.3}    & {\ul 73.7}    & {\ul 78.1}    & {\ul 81.9}    & {\ul 77.8}    & {\ul 74.1}    & {\ul 76.7}    & {\ul 74.4}    & 65.6          & {\ul 68.3}    & 75.6          & {\ul 82.2}    \\
\rowcolor{gray!20}
Ours                        & \textbf{100.0} & \textbf{83.3} & \textbf{90.0} & \textbf{93.3} & \textbf{92.6} & \textbf{92.2} & \textbf{82.9} & \textbf{83.3} & \textbf{88.1} & \textbf{98.1} & \textbf{84.9} & \textbf{85.3} & \textbf{78.7} & \textbf{87.7} \\ \bottomrule
\end{tabular}
}
\resizebox{\linewidth}{!}{
\begin{tabular}{l|cccccccccccccc}
\toprule
Method                      & cap3          & cap4          & cap5          & cup0          & cup1          & eraser0       & headset0      & headset1      & helmet0       & helmet1       & helmet2       & helmet3       & jar0          & micro.        \\ \midrule
BTF (Raw) (CVPR 23')        & 52.7          & 46.8          & 37.3          & 40.3          & 52.1          & 52.5          & 37.8          & 51.5          & 55.3          & 34.9          & 60.2          & 52.6          & 42.0          & 56.3          \\
BTF (FPFH)                  & 52.2          & 52.0          & 58.6          & 58.6          & 61.0          & 71.9          & 52.0          & 49.0          & 57.1          & 71.9          & 54.2          & 44.4          & 42.4          & 67.1          \\
M3DM (CVPR 23')             & 42.3          & {\ul 77.7}    & 63.9          & 53.9          & 55.6          & 62.7          & 57.7          & 61.7          & 52.6          & 42.7          & 62.3          & 37.4          & 44.1          & 35.7          \\
PatchCore (FPFH) (CVPR 22') & 45.3          & 75.7          & \textbf{79.0} & 60.0          & 58.6          & 65.7          & 58.3          & 63.7          & 54.6          & 48.4          & 42.5          & 40.4          & 47.2          & 38.8          \\
PatchCore (PointMAE)        & 47.6          & 72.7          & 53.8          & 61.0          & 55.6          & 67.7          & 59.1          & 62.7          & 55.6          & 55.2          & 44.7          & 42.4          & 48.3          & 48.8          \\
CPMF (PR 24')               & 55.1          & 55.3          & {\ul 69.7}    & 49.7          & 49.9          & 68.9          & 64.3          & 45.8          & 55.5          & 58.9          & 46.2          & 52.0          & 61.0          & 50.9          \\
Reg3D-AD (NeurIPS 23')      & 72.5          & 64.3          & 46.7          & 51.0          & 53.8          & 34.3          & 53.7          & 61.0          & 60.0          & 38.1          & 61.4          & 36.7          & 59.2          & 41.4          \\
IMRNet (CVPR 24')           & {\ul 77.5}    & 65.2          & 65.2          & 64.3          & {\ul 75.7}    & 54.8          & 72.0          & 67.6          & 59.7          & 60.0          & {\ul 64.1}    & 57.3          & 78.0          & 75.5          \\
R3D-AD (ECCV 24')           & 73.0          & 68.1          & 67.0          & {\ul 77.6}    & {\ul 75.7}    & {\ul 89.0}    & {\ul 73.8}    & {\ul 79.5}    & {\ul 75.7}    & {\ul 72.0}    & 63.3          & {\ul 70.7}    & {\ul 83.8}    & {\ul 76.2}    \\
\rowcolor{gray!20}
Ours                        & \textbf{85.9} & \textbf{79.2} & 67.0          & \textbf{87.1} & \textbf{83.3} & \textbf{99.5} & \textbf{80.8} & \textbf{92.3} & \textbf{76.2} & \textbf{96.1} & \textbf{86.9} & \textbf{75.4} & \textbf{86.6} & \textbf{77.6} \\ \bottomrule
\end{tabular}
}
\resizebox{\linewidth}{!}{
\begin{tabular}{l@{\hspace{-1pt}}|cccccccccccc|cc}
\toprule
Method                      & shelf0        & tap0          & tap1          & vase0         & vase1         & vase2         & vase3         & vase4         & vase5         & vase7         & vase8         & vase9         & Average       & Mean rank    \\ \midrule
BTF (Raw) (CVPR 23')        & 16.4          & 52.5          & 57.3          & 53.1          & 54.9          & 41.0          & 71.7          & 42.5          & 58.5          & 44.8          & 42.4          & 56.4          & 49.3          & 7.7          \\
BTF (FPFH)                  & 60.9          & 56.0          & 54.6          & 34.2          & 21.9          & 54.6          & 69.9          & 51.0          & 40.9          & 51.8          & 66.8          & 26.8          & 52.8          & 7.0          \\
M3DM (CVPR 23')             & 56.4          & \textbf{75.4} & 73.9          & 42.3          & 42.7          & 73.7          & 43.9          & 47.6          & 31.7          & 65.7          & 66.3          & 66.3          & 55.2          & 6.8          \\
PatchCore (FPFH) (CVPR 22') & 49.4          & {\ul 75.3}    & {\ul 76.6}    & 45.5          & 42.3          & 72.1          & 44.9          & 50.6          & 41.7          & 69.3          & 66.2          & 66.0          & 56.8          & 6.3          \\
PatchCore (PointMAE)        & 52.3          & 45.8          & 53.8          & 44.7          & 55.2          & 74.1          & 46.0          & 51.6          & 57.9          & 65.0          & 66.3          & 62.9          & 56.2          & 6.4          \\
CPMF (PR 24')               & 68.5          & 35.9          & 69.7          & 45.1          & 34.5          & 58.2          & 58.2          & 51.4          & 61.8          & 39.7          & 52.9          & 60.9          & 55.9          & 6.3          \\
Reg3D-AD (NeurIPS 23')      & {\ul 68.8} & 67.6          & 64.1          & 53.3          & 70.2          & 60.5          & 65.0          & 50.0          & 52.0          & 46.2          & 62.0          & 59.4          & 57.2          & 6.4          \\
IMRNet (CVPR 24')           & 60.3          & 67.6          & 69.6          & 53.3          & \textbf{75.7} & 61.4          & 70.0          & 52.4          & 67.6          & 63.5          & 63.0          & 59.4          & 66.1          & 3.9          \\
R3D-AD (ECCV 24')           & \textbf{69.6} & 73.6          & \textbf{90.0} & {\ul 78.8}    & 72.9          & {\ul 75.2}    & {\ul 74.2}    & {\ul 63.0}    & {\ul 75.7}    & {\ul 77.1}    & {\ul 72.1}    & {\ul 71.8}    & {\ul 74.9}    & {\ul 2.2}    \\
\rowcolor{gray!20}
Ours                        & 57.3          & 74.5          & 68.1          & \textbf{85.8} & {\ul 74.2}    & \textbf{95.2} & \textbf{82.1} & \textbf{67.5} & \textbf{85.2} & \textbf{96.6} & \textbf{73.9} & \textbf{83.0} & \textbf{83.9} & \textbf{1.3} \\
\bottomrule
\end{tabular}
}
\caption{Comparison of object-level AUC-ROC results (\%) of various methods on the Anomaly-ShapeNet dataset. The best result per category is \textbf{bold}, while the second best result is {\ul underlined}. Micro. refers to the microphone0 category. BTF (Raw) refers to that the point coordinates are adopted into the BTF method. PFFH and PointMAE denote utilizing Fast Point Feature Histograms~\cite{rusu2009fast} and ShapeNet~\cite{chang2015shapenet} pre-trained PointMAE~\cite{pang2022masked} as the feature extractor, respectively.}
\label{tab:as-o-aucroc}
\end{table*}

\subsection{Anomaly Score for Inference}
The abnormal level for each point in test data is assessed by its predicted offset. Specifically, the anomaly score of a point is calculated by summing the offset distances along three coordinates (xyz).
The point-level anomaly score function $\phi(p_i)$ is defined as:
\begin{equation}
    \phi(p_i) = \left|o_{i,x}^{pre}\right| + \left|o_{i,y}^{pre}\right| + \left|o_{i,z}^{pre}\right|,
\end{equation}
where $p_i \in P$ and $\{o_{i,x}^{pre}, o_{i,y}^{pre}, o_{i,z}^{pre}\} = o_i^{pre} \in O^{pre}$. According to $\phi(p_i)$, the object-level anomaly score function $\phi(P)$ is obtained by:
\begin{equation}
    \phi(P) = \frac{1}{N}\sum_{i=1}^N\phi(p_i).
\end{equation}
The anomaly scores for normal samples or points are expected to be as small as possible. The greater the anomaly score, the more likely that a sample or point is an anomaly.

\begin{table*}[!t]
\centering
\resizebox{\linewidth}{!}{
\begin{tabular}{l|cccccccccccccc}
\toprule
Method                      & ashtray0      & bag0          & bottle0       & bottle1       & bottle3       & bowl0         & bowl1         & bowl2         & bowl3         & bowl4         & bowl5         & bucket0       & bucket1       & cap0          \\ \midrule
BTF (Raw) (CVPR 23')        & 51.2          & 43.0          & 55.1          & 49.1          & {\ul 72.0}    & 52.4          & 46.4          & 42.6          & {\ul 68.5}    & 56.3          & 51.7          & 61.7          & 68.6          & 52.4          \\
BTF (FPFH)                  & 62.4          & {\ul 74.6}    & 64.1          & 54.9          & 62.2          & 71.0          & {\ul 76.8}    & 51.8          & 59.0          & 67.9          & 69.9          & 40.1          & 63.3          & {\ul 73.0}    \\
M3DM (CVPR 23')             & 57.7          & 63.7          & 66.3          & 63.7          & 53.2          & 65.8          & 66.3          & {\ul 69.4}    & 65.7          & 62.4          & 48.9          & {\ul 69.8}    & 69.9          & 53.1          \\
PatchCore (FPFH) (CVPR 22') & 59.7          & 57.4          & 65.4          & 68.7          & 51.2          & 52.4          & 53.1          & 62.5          & 32.7          & 72.0          & 35.8          & 45.9          & 57.1          & 47.2          \\
PatchCore (PointMAE)        & 49.5          & 67.4          & 55.3          & 60.6          & 65.3          & 52.7          & 52.4          & 51.5          & 58.1          & 50.1          & 56.2          & 58.6          & 57.4          & 54.4          \\
CPMF (PR 24')               & 61.5          & 65.5          & 52.1          & 57.1          & 43.5          & 74.5          & 48.8          & 63.5          & 64.1          & 68.3          & 68.4          & 48.6          & 60.1          & 60.1          \\
Reg3D-AD (NeurIPS 23')      & {\ul 69.8}    & 71.5          & {\ul 88.6}    & 69.6          & 52.5          & 77.5          & 61.5          & 59.3          & 65.4          & {\ul 80.0}    & 69.1          & 61.9          & 75.2          & 63.2          \\
IMRNet (CVPR 24')           & 67.1          & 66.8          & 55.6          & {\ul 70.2}    & 64.1          & {\ul 78.1}    & 70.5          & 68.4          & 59.9          & 57.6          & {\ul 71.5}    & 58.5          & {\ul 77.4}    & 71.5          \\
\rowcolor{gray!20}
Ours                        & \textbf{96.2} & \textbf{94.9} & \textbf{91.2} & \textbf{84.4} & \textbf{88.0} & \textbf{97.8} & \textbf{91.4} & \textbf{91.8} & \textbf{93.5} & \textbf{96.7} & \textbf{94.1} & \textbf{75.5} & \textbf{89.9} & \textbf{95.7} \\ \bottomrule
\end{tabular}
}
\resizebox{\linewidth}{!}{
\begin{tabular}{l|cccccccccccccc}
\toprule
Method                      & cap3          & cap4          & cap5          & cup0          & cup1          & eraser0       & headset0      & headset1      & helmet0       & helmet1       & helmet2       & helmet3       & jar0          & micro.        \\ \midrule
BTF (Raw) (CVPR 23')        & 68.7          & 46.9          & 37.3          & 63.2          & 56.1          & 63.7          & 57.8          & 47.5          & 50.4          & 44.9          & 60.5          & 70.0          & 42.3          & 58.3          \\
BTF (FPFH)                  & 65.8          & 52.4          & 58.6          & {\ul 79.0}    & 61.9          & 71.9          & 62.0          & 59.1          & 57.5          & {\ul 74.9}    & 64.3          & 72.4          & 42.7          & 67.5          \\
M3DM (CVPR 23')             & 60.5          & 71.8          & 65.5          & 71.5          & 55.6          & 71.0          & 58.1          & 58.5          & 59.9          & 42.7          & 62.3          & 65.5          & 54.1          & 35.8          \\
PatchCore (FPFH) (CVPR 22') & 65.3          & 59.5          & {\ul 79.5}    & 65.5          & 59.6          & {\ul 81.0}    & 58.3          & 46.4          & 54.8          & 48.9          & 45.5          & {\ul 73.7}    & 47.8          & 48.8          \\
PatchCore (PointMAE)        & 48.8          & 72.5          & 54.5          & 51.0          & {\ul 85.6}    & 37.8          & 57.5          & 42.3          & 58.0          & 56.2          & 65.1          & 61.5          & 48.7          & \textbf{88.6} \\
CPMF (PR 24')               & 55.1          & 55.3          & {\ul 55.1}    & 49.7          & 50.9          & 68.9          & 69.9          & 45.8          & 55.5          & 54.2          & 51.5          & 52.0          & 61.1          & 54.5          \\
Reg3D-AD (NeurIPS 23')      & {\ul 71.8}    & {\ul 81.5}    & 46.7          & 68.5          & 69.8          & 75.5          & 58.0          & {\ul 62.6}    & {\ul 60.0}    & 62.4          & {\ul 82.5} & 62.0          & 59.9          & 59.9          \\
IMRNet (CVPR 24')           & 70.6          & 75.3          & 74.2          & 64.3          & 68.8          & 54.8          & {\ul 70.5}    & 47.6          & 59.8          & 60.4          & 64.4          & 66.3          & {\ul 76.5}    & 74.2          \\
\rowcolor{gray!20}
Ours                        & \textbf{94.8} & \textbf{94.0} & \textbf{86.4} & \textbf{90.9} & \textbf{93.2} & \textbf{97.4} & \textbf{82.3} & \textbf{90.7} & \textbf{87.8} & \textbf{94.8} & \textbf{93.2} & \textbf{84.6} & \textbf{87.1} & {\ul 81.0}    \\ \bottomrule
\end{tabular}
}
\resizebox{\linewidth}{!}{
\begin{tabular}{l@{\hspace{-1pt}}|cccccccccccc|cc}
\toprule
Method                      & shelf0        & tap0          & tap1          & vase0         & vase1         & vase2         & vase3         & vase4         & vase5         & vase7         & vase8         & vase9         & Average       & Mean rank    \\ \midrule
BTF (Raw) (CVPR 23')        & 46.4          & 52.7          & 56.4          & 61.8          & 54.9          & 40.3          & 60.2          & 61.3          & 58.5          & 57.8          & 55.0          & 56.4          & 55.0          & 6.9          \\
BTF (FPFH)                  & 61.9          & 56.8          & 59.6          & 64.2          & 61.9          & 64.6          & {\ul 69.9}    & 71.0          & 42.9          & 54.0          & 66.2          & 56.8          & 62.8          & 4.8          \\
M3DM (CVPR 23')             & 55.4          & 65.4          & 71.2          & 60.8          & 60.2          & 73.7          & 65.8          & 65.5          & 64.2          & 51.7          & 55.1          & 66.3          & 61.6          & 5.1          \\
PatchCore (FPFH) (CVPR 22') & 61.3          & 73.3          & \textbf{76.8} & 65.5          & 45.3          & 72.1          & 43.0          & 50.5          & 44.7          & 69.3          & 57.5          & 66.3          & 58.0          & 5.9          \\
PatchCore (PointMAE)        & 54.3          & \textbf{85.8} & 54.1          & {\ul 67.7}    & 55.1          & {\ul 74.2}    & 46.5          & 52.3          & 57.2          & 65.1          & 36.4          & 42.3          & 57.7          & 6.2          \\
CPMF (PR 24')               & \textbf{78.3} & 45.8          & 65.7          & 45.8          & 48.6          & 58.2          & 58.2          & 51.4          & 65.1          & 50.4          & 52.9          & 54.5          & 57.3          & 6.5          \\
Reg3D-AD (NeurIPS 23')      & {\ul 68.8}    & 58.9          & {\ul 74.1}    & 54.8          & 60.2          & 40.5          & 51.1          & {\ul 75.5}    & 62.4          & {\ul 88.1}    & {\ul 81.1}    & {\ul 69.4}    & {\ul 66.8}    & {\ul 3.8}    \\
IMRNet (CVPR 24')           & 60.5          & 68.1          & 69.9          & 53.5          & {\ul 68.5}    & 61.4          & 40.1          & 52.4          & {\ul 68.2}    & 59.3          & 63.5          & 69.1          & 65.0          & 4.2          \\
\rowcolor{gray!20}
Ours                        & 66.3          & {\ul 78.3}    & 69.2          & \textbf{95.5} & \textbf{88.2} & \textbf{97.8} & \textbf{88.4} & \textbf{90.2} & \textbf{93.7} & \textbf{98.2} & \textbf{95.0} & \textbf{95.2} & \textbf{89.8} & \textbf{1.2} \\ \bottomrule
\end{tabular}
}
\caption{Comparsion of point-level AUC-ROC results on the Anomaly-ShapeNet dataset.}
\label{tab:as-p-aucroc}
\end{table*}

\section{Experiments}
\subsection{Experimental Settings}
\textbf{Datasets.} Following two benchmarks~\cite{liu2023real3d,li2024towards}, our evaluation encompasses two existing pure point cloud anomaly detection datasets: Anomaly-ShapeNet~\cite{li2024towards} and Real3D-AD~\cite{liu2023real3d}. Anomaly-ShapeNet is a synthesis dataset based on ShapeNet~\cite{chang2015shapenet} dataset. It consists of 1,600 samples belonging to 40 categories. The training set of each category contains 4 normal samples. Real3D-AD is a high-resolution point cloud dataset based on real objects of 12 categories. Each category contains 4 training normal samples and 100 test instances. There is a large difference between training and test samples in the Real3D-AD dataset where training samples undergo 360$^\circ$ scan, while test samples are scanned on only one side.

\textbf{Evaluation metrics.} Experiments are conducted by following previous work~\cite{liu2023real3d, li2024towards}. Area Under the Receiver-Operating-Characteristic Curve (AUC-ROC) is utilized as our evaluation criterion. It can objectively evaluate detection (object-level) and localization (point-level) performance without making any assumption on the decision threshold.

\subsection{Implementation Details}
MinkUNet34C~\cite{choy2019fully,choy20194d} serves as our backbone for feature extraction. A three-layer MLP with PReLU activation function forms the offset predictor.
We set the dimension of latent features $C$ to 32, and the voxel size to 0.03. Our network is trained for 1,000 epochs with a batch size of 32 (the training set is replicated 100 times to obtain 400 samples). The model parameters are optimized by Adam with an initial learning rate of 0.001, which decays with the cosine anneal schedule~\cite{loshchilov2017sgdr}. Our method does not involve point cloud downsampling. Training samples are applied with random rotation before normalization. All input point clouds are normalized by aligning their center of gravity with the origin of coordinates and scaling their dimensions to range from -1 to 1.
We set the patch number $J$ to 64 for our Norm-AS, which is performed after normalization. The \textit{normal vectors} are obtained from official dataset OBJ files.

\subsection{Baseline Methods}
The proposed method is compared with eight outstanding methods: BTF~\cite{horwitz2023back}, M3DM~\cite{wang2023multimodal}, PatchCore~\cite{roth2022towards}, CPMF~\cite{cao2024complementary}, Reg3D-AD \cite{liu2023real3d}, IMRNet~\cite{li2024towards}, R3D-AD~\cite{zhou2024r3dad}, and Group3AD~\cite{zhu2024towards}. PatchCore is originally a 2D anomaly detection method and is applied to 3D by replacing feature extractors. The results of BTF, M3DM, PatchCore, and CPMF are implemented by Real3D-AD and IMRNet. The results of other methods are obtained from their papers.

\begin{table*}[!t]
\centering
\resizebox{\linewidth}{!}{
\begin{tabular}{@{}l|cccccccccc>{\columncolor{gray!20}}c}
\toprule
Category    & \begin{tabular}[c]{@{}c@{}}BTF\\(Raw)\\(CVPR 23')\end{tabular} & \begin{tabular}[c]{@{}c@{}}BTF\\(FPFH)\end{tabular} & \begin{tabular}[c]{@{}c@{}}M3DM\\ (CVPR 23') \end{tabular}  & \begin{tabular}[c]{@{}c@{}}PatchCore\\(FPFH)\\ (CVPR 22')\end{tabular} & \begin{tabular}[c]{@{}c@{}}PatchCore\\(PointMAE)\end{tabular} & \begin{tabular}[c]{@{}c@{}}CPMF\\ (PR 24') \end{tabular}       & \begin{tabular}[c]{@{}c@{}}Reg3D-AD\\ (NeurIPS 23')\end{tabular}   & \begin{tabular}[c]{@{}c@{}}IMRNet\\ (CVPR 24')\end{tabular}        & \begin{tabular}[c]{@{}c@{}}R3D-AD\\ (ECCV 24')\end{tabular} & \begin{tabular}[c]{@{}c@{}}Group3AD\\ (MM 24')\end{tabular}        & Ours          \\ \midrule
Airplane & 73.0                                                & 52.0                                                 & 43.4 & \textbf{88.2}                                              & 72.6                                                           & 70.1 & 71.6          & 76.2       & 77.2          & 74.4          & {\ul 80.4}    \\
Car      & 64.7                                                & 56.0                                                 & 54.1 & 59.0                                                       & 49.8                                                           & 55.1 & 69.7          & {\ul 71.1} & 69.3          & \textbf{72.8} & 65.4          \\
Candy    & 53.9                                                & 63.0                                                 & 55.2 & 54.1                                                       & 66.3                                                           & 55.2 & 68.5          & 75.5       & 71.3          & \textbf{84.7} & {\ul 78.5}    \\
Chicken  & 78.9                                                & 43.2                                                 & 68.3 & {\ul 83.7}                                                 & 82.7                                                           & 50.4 & \textbf{85.2} & 78.0       & 71.4          & 78.6          & 68.6          \\
Diamond  & 70.7                                                & 54.5                                                 & 60.2 & 57.4                                                       & 78.3                                                           & 52.3 & 90.0          & {\ul 90.5} & 68.5          & \textbf{93.2} & 80.1          \\
Duck     & 69.1                                                & 78.4                                                 & 43.3 & 54.6                                                       & 48.9                                                           & 58.2 & 58.4          & 51.7       & \textbf{90.9} & 67.9          & {\ul 82.0}    \\
Fish     & 60.2                                                & 54.9                                                 & 54.0 & 67.5                                                       & 63.0                                                           & 55.8 & {\ul 91.5}    & 88.0       & 69.2          & \textbf{97.6} & 85.9          \\
Gemstone & {\ul 68.6}                                          & 64.8                                                 & 64.4 & 37.0                                                       & 37.4                                                           & 58.9 & 41.7          & 67.4       & 66.5          & 53.9          & \textbf{69.3} \\
Seahorse & 59.6                                                & {\ul 77.9}                                           & 49.5 & 50.5                                                       & 53.9                                                           & 72.9 & 76.2          & 60.4       & 72.0          & \textbf{84.1} & 75.6          \\
Shell    & 39.6                                                & 75.4                                                 & 69.4 & 58.9                                                       & 50.1                                                           & 65.3 & 58.3          & 66.5       & \textbf{84.0} & 58.5          & {\ul 80.0}    \\
Starfish & 53.0                                                & 57.5                                                 & 55.1 & 44.1                                                       & 51.9                                                           & 70.0 & 50.6          & 67.4       & {\ul 70.1}    & 56.2          & \textbf{75.8} \\
Toffees  & 70.3                                                & 46.2                                                 & 45.0 & 56.5                                                       & 58.5                                                           & 39.0 & \textbf{82.7} & 77.4       & 70.3          & {\ul 79.6}    & 77.1          \\ \midrule
Average  & 63.5                                                & 60.3                                                 & 55.2 & 59.3                                                       & 59.4                                                           & 58.6 & 70.4          & 72.5       & 73.4          & {\ul 75.1}    & \textbf{76.5} \\
Mean rank & 6.5                                                 & 6.9                                                  & 8.8  & 7.5                                                        & 7.8                                                            & 7.9  & 5.0           & 4.2        & 4.0           & {\ul 3.6}     & \textbf{3.2}  \\ \bottomrule
\end{tabular}}
\caption{Object-level AUC-ROC results of our method and competitors on the Real3D-AD dataset.}
\label{tab:r-o-aucroc}
\end{table*}

\begin{table}[!t]
\centering
\resizebox{\linewidth}{!}{
\begin{tabular}{@{}c|ccc>{\columncolor{gray!20}}c}
\toprule
Method               & Variant 1 & Variant 2 & Variant 3  & Ours          \\ \midrule
$\mathcal{L}_{dist}$              & \checkmark          & -         & \checkmark           & \checkmark              \\
$\mathcal{L}_{dir}$               & -         & \checkmark          & \checkmark           & \checkmark              \\
\textit{Normal vector}        & \checkmark          & \checkmark          & -          & \checkmark              \\ \midrule
Object-level AUC-ROC & 50.3      & 67.5      & {\ul 81.1} & \textbf{84.2} \\
Point-level AUC-ROC  & 50.4      & 74.9      & {\ul 78.4} & \textbf{87.8} \\ \bottomrule
\end{tabular}}
\caption{Ablation study of our method and its variants.}
\label{tab:ablation}
\end{table}

\subsection{Main Results}
\subsubsection{Results on Anomaly-ShapeNet}
Table~\ref{tab:as-o-aucroc} and \ref{tab:as-p-aucroc} respectively present the results for detection and localization, comparing our method against competing methods on the Anomaly-ShapeNet dataset.
Our method achieves the best overall performance on both tasks, outperforming the second-best method by an average of $9.0\%$ on detection and $23.0\%$ on localization. To prevent a few categories from dominating the averaged results, we also calculate the mean rank ($\downarrow$) for comparison. Notably, our method obtains the best (lowest) mean rank on both object-level and point-level AUC-ROC, significantly surpassing the competing methods.
At the category level, our approach not only outperforms competitors in the majority of categories, but also maintains competitive performance in the remaining ones.
Additionally, our method attains considerable performance gains compared to the best contestant on various categories, such as bag0 and bowl4.
Generally, these comparison results validate the superiority of our method. Furthermore, object-level AUC-PR results can be found in Table~\ref{tab:as-o-ap} of Supplementary~\ref{sec:o-as-ap}.

\subsubsection{Results on Real3D-AD}
Table~\ref{tab:r-o-aucroc} depicts the comparison of object-level AUC-ROC results on the Real3D-AD dataset. According to the mean rank, our method secures the first place by a narrow margin, with an average AUC-ROC improvement of 1.4\% over the second-best method. At the category level, our method achieves the best or the second-best results in 6 categories and exhibits commendable performance in the rest. 
It is noted that there is a huge gap between training data and test data of the Real3D-AD dataset, i.e., training samples are scanned 360$^\circ$, but test point clouds are scanned only on one side.
The memory bank-based methods (Reg3D-AD, Group3AD) have an advantage when dealing with such situations, as they leverage the technique of template registration to detect anomalies.
Despite this, our method still surpasses them on both average performance and mean rank. Compared to reconstruction-based methods, our method achieves the best results in most categories: 8 compared to R3D-AD and 7 compared to IMRNet. Overall, these comparison results evidences the effectiveness of our method.

\subsection{Ablation Study}
\label{sec:ablation}
Fifteen categories ending in 0 of the Anomaly-ShapeNet dataset are selected to conduct the ablation study. The averaged results are reported in Table~\ref{tab:ablation}.

\textbf{Normal representation learning heavily relies on $\mathcal{L}_{dir}$:} We design ``Variant 1", where the model is supervised solely by $\mathcal{L}_{dist}$. The absence of $\mathcal{L}_{dir}$ causes the network to struggle with precisely estimating the offset direction of pseudo-abnormal points.
According to the experimental results, the performance of ``Variant" is much lower than that of our method, validating the significance of $\mathcal{L}_{dir}$ for capturing effective normal representations.

\textbf{$\mathcal{L}_{dist}$ is essential for capturing effective normal representations:} ``Variant 2" learns a single objective of predicting point offset direction. Evidently, it is significantly inferior to our method. Without $L_{dist}$, the model fails to learn offset distance for both normal and pseudo-abnormal points. Additionally, it completely disregards normal points as $\mathcal{L}_{dir}$ is not applicable for them. Therefore, $\mathcal{L}_{dist}$ is indispensable in our offset prediction-based framework.

\begin{figure}[!t]
    \centering
    \includegraphics[width=0.9\linewidth]{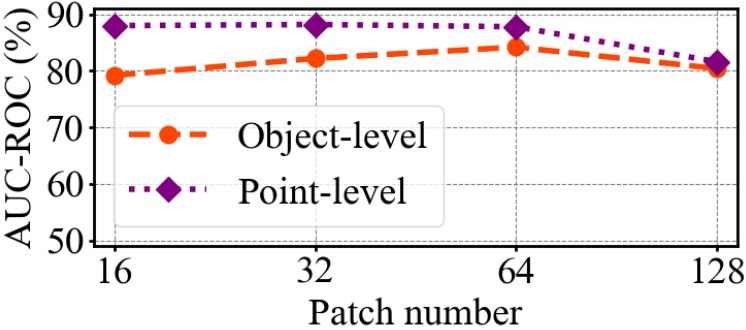}
    \caption{Detection and localization performance \textit{vs.} patch no.}
    \label{fig:sensitivity}
\end{figure}

\begin{figure*}[!t]
    \centering
    \includegraphics[width=\linewidth]{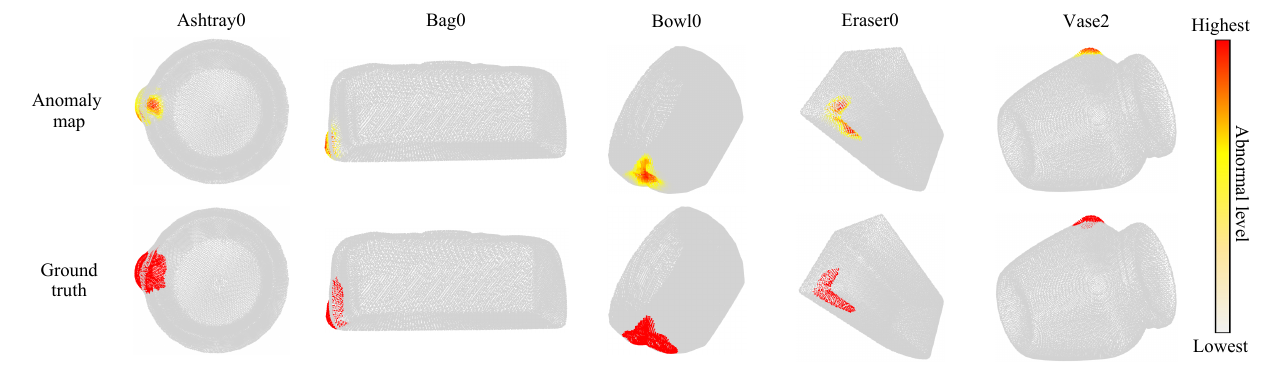}
    \caption{Qualitative results of localization on five categories of the Anomaly-ShapeNet dataset, where brighter color refers to a higher abnormal level.}
    \label{fig:localization}
\end{figure*}

\begin{figure}[!t]
    \centering
    \includegraphics[width=\linewidth]{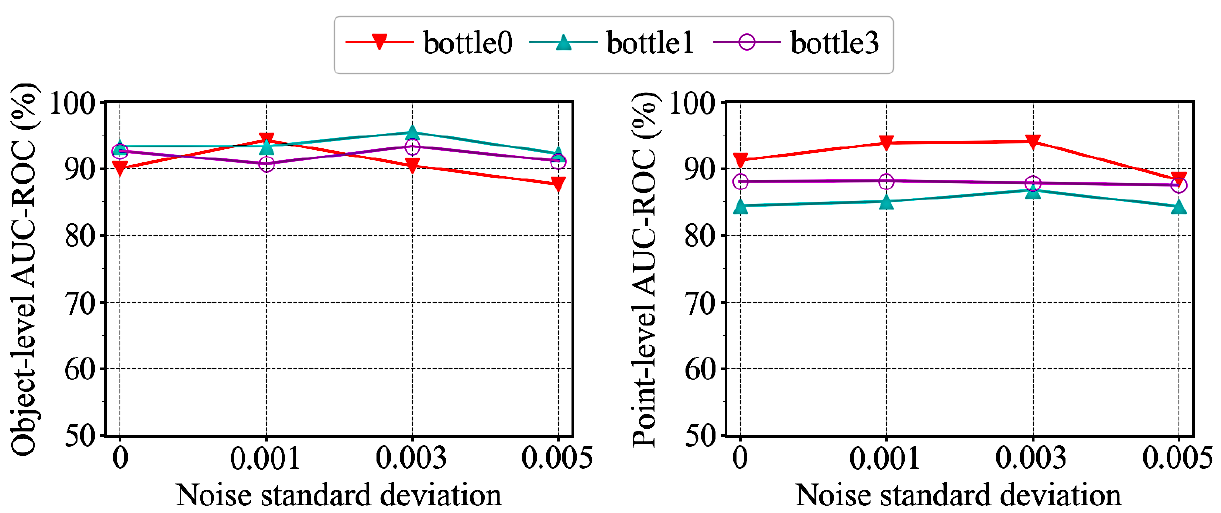}
    \caption{Detection and localization performance \textit{vs.} noise with various standard deviations.}
    \label{fig:robustness}
\end{figure}

\textbf{Generating pseudo anomalies guided by \textit{normal vectors} helps the normal representation learning:} A substantial performance drop is observed in ``Variant 3", since moving points in random directions may produce unsuitable pseudo anomalies that confuse the model, resulting in less efficient learning. This indicates that the proposed Norm-AS is crucial for facilitating the extraction of normal representations. Besides, the performance of ``Variant 3" further demonstrates the superiority of our offset prediction framework compared to reconstruction-based R3D-AD (77.2\%).

\subsection{Analysis on Patch Number}
\label{sec:sen}

Figure~\ref{fig:sensitivity} reports the object-level and point-level AUC-ROC results \textit{vs.} different patch numbers, which are average on fifteen categories ending in 0 of the Anomaly-ShapeNet dataset. The size of pseudo-abnormal regions is inversely correlated with the patch number $J$. An appropriate size is crucial for learning normal representations. Difficulty in predicting point offset for a region that is too large may hinder the model's convergence. Conversely, learning point offsets for a region that is too small may prevent the model from capturing sufficient normal representations.
However, despite these effects, our method is generally less sensitive to the size of pseudo-abnormal regions. According to the presented results, the detection and localization performance reach their best when the patch numbers are 32 and 64, respectively. We set the patch number to 64 in our implementation to achieve the best detection performance, at the cost of a slight sacrifice in localization performance.

\subsection{Robustness to Noisy Data}
In real-world scenarios, the complexity of environments and the instability of equipment may result in scanned point clouds containing noise, i.e., noisy data.
To analyze the robustness of our method concerning noisy data, we conduct experiments on test samples containing Gaussian noise with a standard deviation of 0, 0.001, 0.003, and 0.005 (0 denotes clean data).

Selecting bottle0, 1, and 3 as illustrative categories, analysis results are presented in Figure~\ref{fig:robustness}.
It is observed that performance only drops slightly as the noise standard deviation increases. Additionally, the worst case of our method is still higher than competing methods tested on clean data (such as 73.3\%, 73.7\%, and 78.1\% object-level AUC-ROC of R3D-AD on bottle0, 1, and 3). Such empirical results evidence the robustness of our method to noisy data. We visualize noisy point clouds in Figure~\ref{fig:noisydata} of Supplementary~\ref{sec:noisydata}.

\subsection{Qualitative Results}
Figure~\ref{fig:localization} illustrates anomaly maps for localization on five categories of the Anomaly-ShapeNet dataset. The anomaly map is directly obtained by performing the point-level scoring function $\phi(p_i)$ during inference. Our method precisely locates the abnormal regions, and also assigns relatively much lower abnormal levels to normal points. This validates the effectiveness of our method.

\section{Conclusion}
In this paper, we design a novel framework PO3AD based on point offset prediction to capture effective normal representations for 3D point cloud anomaly detection.
Moreover, we propose an anomaly simulation method named Norm-AS guided by \textit{normal vectors}, creating credible pseudo anomalies from normal samples to facilitate the distillation of normal representations. Extensive experiments conducted on the Anomaly-ShapeNet and Real3D-AD datasets evidence that our method outperforms the existing best methods.

\textbf{Limitations and future work.} It is imperative to note that our current design is still under the one-model-per-category learning paradigm, i.e., each category needs a specifically trained detection model, leading to prohibitive computational and storage. In future work, we intend to investigate the inter-category common patterns to explore a one-model-all-category learning paradigm for point cloud anomaly detection.

{
    \small
    \bibliographystyle{ieeenat_fullname}
    \bibliography{ref}
}

\clearpage
\setcounter{page}{1}
\maketitlesupplementary

\begin{figure*}[!hb]
    \centering
    \includegraphics[width=0.9\linewidth]{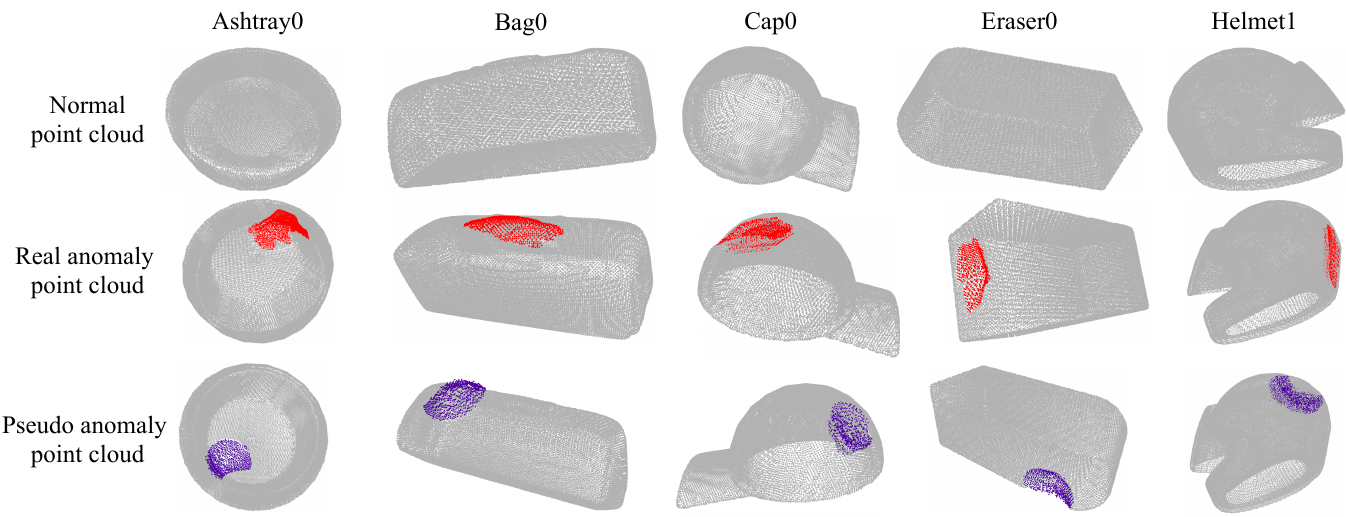}
    \caption{Visualizations of normal, real anomaly, and our pseudo anomaly samples.}
    \label{fig:pseudovis}
\end{figure*}

\begin{figure*}[!hb]
    \centering
    \includegraphics[width=0.9\linewidth]{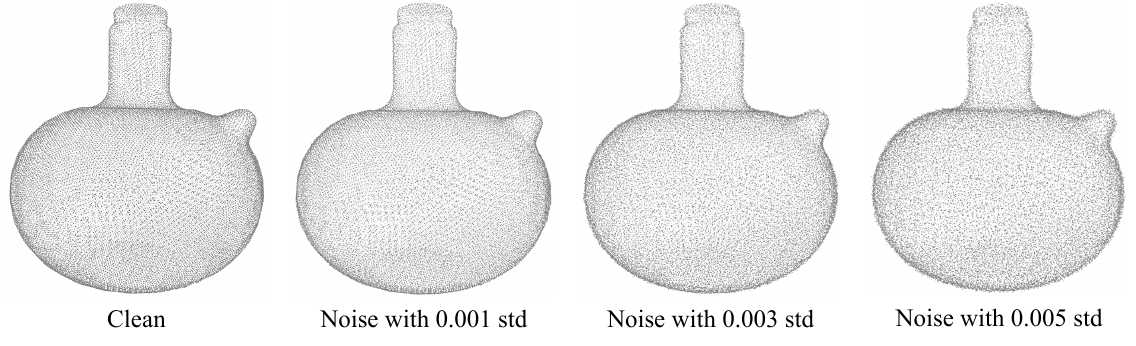}
    \caption{Visualizations of clean, and noisy point clouds with various standard deviations (std).}
    \label{fig:noisydata}
\end{figure*}

\begin{figure*}[!hb]
    \centering
    \includegraphics[width=0.9\linewidth]{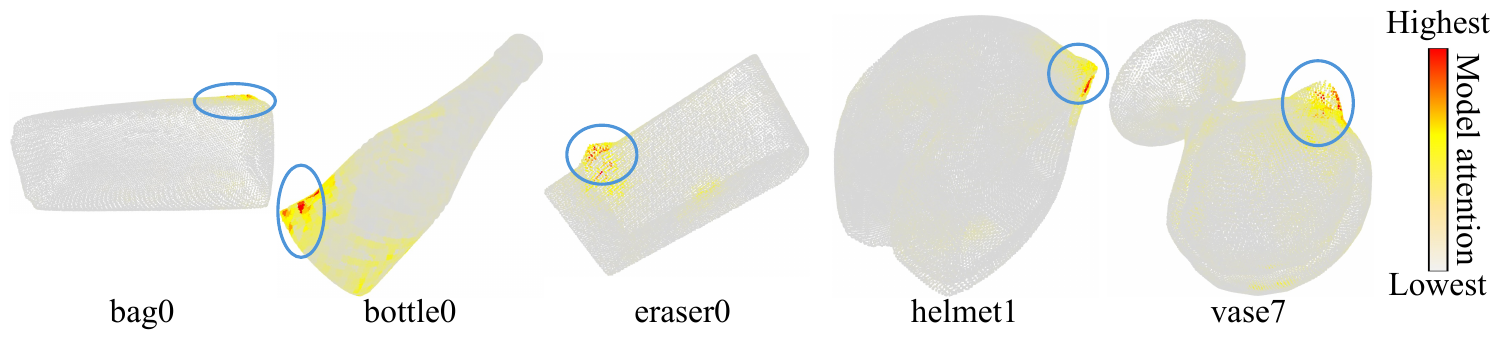}
    \caption{Visualization of model attention maps, pseudo-abnormal regions are marked with blue circles.}
    \label{fig:supp_attn}
\end{figure*}

\section{Visualizations of Our Pseudo Anomalies}
\label{sec:pseudovis}
Fig.~\ref{fig:pseudovis} presents visualizations of normal, real anomaly, and our pseudo anomaly samples. It is observed that the pseudo anomalies generated using the Norm-AS method closely resemble real anomalies, thereby supporting the efficacy of our approach in creating credible pseudo anomaly instances.

\section{Additional Experimental Results}
\label{sec:o-as-ap}
\textbf{Object-level AUC-PR results on Anomaly-ShapeNet.}
Table~\ref{tab:as-o-ap} presents the results of our comparative analysis of object-level AUC-PR on the Anomaly-ShapeNet dataset. 
The results indicate that our method achieves the best mean rank and significantly outperforms the second-best method by an average of 26.0\% AUC-PR. Such experimental results evidence the superiority of our method.

\textbf{Results of using PTv3 [\textcolor{cvprblue}{42}] as backbone.} Four categories with significant performance improvement are selected to conduct experiments using PTv3 as the backbone, results are presented in Table~\ref{tab:backbone}. Applying PTv3 as the backbone for feature extraction, our method still outperforms R3D-AD, validating the effectiveness of our offset prediction strategy.

\section{Visualizations of Noisy Data}
\label{sec:noisydata}
Figure~\ref{fig:noisydata} depicts the visualizations of a clean point cloud and its noisy variants with various standard deviations.
It is observed that as the noise standard deviation grows, the point cloud surface becomes progressively less smooth.

\section{Visualizations of Model Attention Maps}
Visualizations of model attention maps are presented in Figure~\ref{fig:supp_attn}. Evidently, our method successfully focuses on pseudo-abnormal regions, allowing the effective extraction of normal representations for anomaly detection.

\begin{table*}[!hb]
\centering
\resizebox{\linewidth}{!}{
\begin{tabular}{l|cccccccccccccc}
\toprule
Method                      & ashtray0      & bag0          & bottle0       & bottle1       & bottle3       & bowl0         & bowl1         & bowl2         & bowl3         & bowl4         & bowl5         & bucket0       & bucket1       & cap0          \\ \midrule
BTF (Raw) (CVPR 23')        & 57.8          & 45.8          & 46.6          & 57.3          & 54.3          & 58.8          & 46.4          & 57.6          & {\ul 65.4}    & 60.1          & 61.5          & 65.2          & 62.0          & 65.9          \\
BTF (FPFH)                  & 65.1          & 55.1          & 64.4          & 62.5          & 60.2          & 57.6          & {\ul 64.8}    & 51.5          & 49.9          & 63.2          & {\ul 69.9}    & 48.3          & 64.8          & 61.8          \\
M3DM (CVPR 23')             & 63.2          & 64.2          & {\ul 76.3}    & 67.4          & 45.1          & 52.5          & 51.5          & 63.0          & 63.5          & 57.1          & 60.1          & 60.9          & 50.7          & 56.4          \\
PatchCore (FPFH) (CVPR 22') & 44.5          & 60.8          & 61.5          & 67.7          & 57.9          & 54.8          & 54.5          & 61.1          & 62.0          & 57.5          & 54.1          & 60.4          & 56.5          & 58.5          \\
PatchCore (PointMAE)        & {\ul 67.9}    & 60.1          & 54.5          & 64.5          & {\ul 65.1}    & 56.2          & 61.1          & 45.6          & 55.6          & 60.1          & 58.5          & 54.1          & 64.2          & 56.1          \\
CPMF (PR 24')               & 45.3          & 65.5          & 58.8          & 59.2          & 50.5          & {\ul 77.5}    & 62.1          & 60.1          & 41.8          & {\ul 68.3}    & 68.5          & {\ul 66.2}    & 50.1          & 60.1          \\
Reg3D-AD (NeurIPS 23')      & 58.8          & 60.8          & 63.2          & 69.5          & 47.4          & 49.4          & 51.5          & 49.5          & 44.1          & 62.4          & 55.5          & 63.2          & 71.4          & 69.3          \\
IMRNet (CVPR 24')           & 61.2          & {\ul 66.5}    & 55.8          & {\ul 70.2}    & 64.8          & 48.1          & 50.4          & {\ul 68.1}    & 61.4          & 63.0          & 65.2          & 57.8          & {\ul 73.2}    & {\ul 71.1}    \\
\rowcolor{gray!20}
Ours                        & \textbf{99.9} & \textbf{80.9} & \textbf{92.7} & \textbf{95.9} & \textbf{96.2} & \textbf{94.6} & \textbf{90.5} & \textbf{88.8} & \textbf{92.7} & \textbf{98.5} & \textbf{90.4} & \textbf{92.3} & \textbf{88.2} & \textbf{84.1} \\ \bottomrule
\end{tabular}
}
\resizebox{\linewidth}{!}{
\begin{tabular}{l|cccccccccccccc}
\toprule
Method                      & cap3          & cap4          & cap5          & cup0          & cup1          & eraser0       & headset0      & headset1      & helmet0       & helmet1       & helmet2       & helmet3       & jar0          & micro.        \\ \midrule
BTF (Raw) (CVPR 23')        & 61.2          & 51.5          & 65.3          & 60.1          & 70.1          & 42.5          & 37.9          & 51.5          & 55.9          & 38.8          & 61.5          & 52.6          & 42.8          & 61.3          \\
BTF (FPFH)                  & 57.9          & 54.5          & 59.3          & 58.5          & 65.1          & 71.9          & 53.1          & 52.3          & 56.8          & {\ul 72.1}    & 58.8          & 56.4          & 47.9          & {\ul 66.2}    \\
M3DM (CVPR 23')             & 65.2          & 47.7          & 64.2          & 57.0          & {\ul 75.2}    & 62.5          & 63.2          & 62.3          & 52.8          & 62.7          & {\ul 63.6}    & 45.8          & 55.5          & 46.4          \\
PatchCore (FPFH) (CVPR 22') & 45.7          & 65.5          & 72.5          & 60.4          & 58.6          & 58.4          & {\ul 70.1}    & 60.1          & 52.5          & 63.0          & 47.5          & 49.4          & 49.9          & 33.2          \\
PatchCore (PointMAE)        & 58.3          & {\ul 72.1}    & 54.2          & 64.2          & 71.0          & {\ul 80.1}    & 51.5          & 42.3          & 63.3          & 57.1          & 49.6          & 61.1          & 46.3          & 65.2          \\
CPMF (PR 24')               & 54.1          & 64.5          & 69.7          & {\ul 64.7}    & 60.9          & 54.4          & 60.2          & 61.9          & 33.3          & 50.1          & 47.7          & {\ul 64.5}    & 61.8          & 65.5          \\
Reg3D-AD (NeurIPS 23')      & {\ul 71.1}    & 62.3          & {\ul 77.0}    & 53.1          & 63.8          & 42.4          & 53.8          & 61.7          & 60.0          & 38.1          & \textbf{61.8} & 46.8          & 60.1          & 61.4          \\
IMRNet (CVPR 24')           & 70.2          & 65.8          & 50.2          & 45.5          & 62.7          & 59.9          & {\ul 70.1}    & {\ul 65.6}    & {\ul 69.7}    & 61.5          & 60.2          & 57.5          & {\ul 76.0}    & 55.2          \\
\rowcolor{gray!20}
Ours                        & \textbf{90.6} & \textbf{87.6} & \textbf{80.1} & \textbf{87.9} & \textbf{87.0} & \textbf{99.5} & \textbf{76.5} & \textbf{91.4} & \textbf{86.4} & \textbf{96.1} & \textbf{93.4} & \textbf{84.9} & \textbf{91.5} & \textbf{80.3} \\ \bottomrule
\end{tabular}
}
\resizebox{\linewidth}{!}{
\begin{tabular}{l@{\hspace{-1pt}}|cccccccccccc|cc}
\toprule
Method                      & shelf0        & tap0          & tap1          & vase0         & vase1         & vase2         & vase3         & vase4         & vase5         & vase7         & vase8         & vase9         & Average       & Mean rank    \\ \midrule
BTF (Raw) (CVPR 23')        & 62.4          & 53.5          & 59.4          & 56.2          & 44.1          & 41.3          & {\ul 71.7}    & 42.8          & 61.5          & 54.7          & 41.6          & 48.2          & 54.9          & 6.5          \\
BTF (FPFH)                  & 61.1          & 61.0          & 57.5          & 64.1          & 65.5          & 56.9          & 65.2          & 58.7          & 47.2          & 59.2          & 62.4          & 63.8          & 59.8          & 5.3          \\
M3DM (CVPR 23')             & 66.5          & {\ul 72.2}    & 63.8          & \textbf{78.8} & 65.2          & 61.5          & 55.1          & 52.6          & 63.3          & 64.8          & 46.3          & 65.1          & 60.3          & 5.1          \\
PatchCore (FPFH) (CVPR 22') & 50.4          & 71.2          & 68.4          & 64.5          & 62.3          & {\ul 80.1}    & 48.1          & {\ul 77.7}    & 51.5          & 62.1          & 51.5          & {\ul 66.0}    & 58.8          & 5.6          \\
PatchCore (PointMAE)        & 54.3          & 71.2          & 54.2          & 54.8          & 57.2          & 71.1          & 45.5          & 58.6          & 58.5          & {\ul 65.2}    & 65.5          & 63.4          & 59.5          & 5.6          \\
CPMF (PR 24')               & \textbf{68.1} & 63.9          & 69.7          & 63.2          & 64.5          & 63.2          & 58.8          & 65.5          & 51.8          & 43.2          & {\ul 67.3}    & 61.8          & 59.7          & 5.1          \\
Reg3D-AD (NeurIPS 23')      & 67.5          & 67.6          & 59.9          & 61.5          & 46.8          & 64.1          & 65.1          & 50.5          & 58.8          & 45.5          & 62.9          & 57.4          & 58.4          & 5.6          \\
IMRNet (CVPR 24')           & 62.5          & 40.1          & \textbf{79.6} & 57.3          & {\ul 72.5}    & 65.5          & 70.8          & 52.8          & {\ul 65.4}    & 60.1          & 63.9          & 46.2          & {\ul 62.1}    & {\ul 4.6}    \\
\rowcolor{gray!20}
Ours                        & {\ul 68.0}    & \textbf{85.6} & {\ul 70.9}    & {\ul 75.3}    & \textbf{78.9} & \textbf{96.3} & \textbf{90.2} & \textbf{82.4} & \textbf{87.9} & \textbf{97.1} & \textbf{83.3} & \textbf{90.4} & \textbf{88.1} & \textbf{1.0} \\ \bottomrule
\end{tabular}
}
\caption{Comparison of object-level AUC-PR results on the Anomaly-ShapeNet dataset.}
\label{tab:as-o-ap}
\end{table*}

\begin{table}[H]
\centering
\begin{tabular}{@{}l|cccc@{}}
\toprule
Method                & bottle0 & bowl0 & cap0 & vase0 \\ \midrule
Ours (PointFormer v3) & {\ul 83.3}    & {\ul 85.5}  & {\ul 84.0} & \textbf{87.0}  \\
Ours (MinkUNet34C)    & \textbf{90.0}    & \textbf{92.2}  & \textbf{87.7} & {\ul 85.8}  \\ \midrule
R3D-AD (ECCV 24')     & 73.3    & 81.9  & 82.2 & 78.8  \\ \bottomrule
\end{tabular}
\caption{Object-level AUC-ROC results \textit{vs.} backbone. The last row presents the results of R3D-AD.}
\label{tab:backbone}
\end{table}

\end{document}